# Evolving Controllers for Simulated Car Racing

Julian Togelius and Simon M. Lucas
Department of Computer Science
University of Essex
Colchester CO4 3SQ, United Kingdom
julian@togelius.com, sml@essex.ac.uk

**Abstract-**

This paper describes the evolution of controllers for racing a simulated radio-controlled car around a track, modelled on a real physical track. Five different controller architectures were compared, based on neural networks, force fields and action sequences. The controllers use either egocentric (first person), Newtonian (third person) or no information about the state of the car (open-loop controller). The only controller that able to evolve good racing behaviour was based on a neural network acting on egocentric inputs.

## 1 Introduction

That car racing is a challenging problem, generating considerable public excitement, is evident from the huge amount of time and money invested both in practising and watching physical car racing, and in developing and playing racing games. For the same reasons, the problem(s) cannot sensibly be considered "trivial" or "solved" - no one would want to watch a race where the drivers were perfect.

Though experiments with neural and evolutionary methods have undoubtedly taken place in commercial game studios, these have not been published for reasons of commercial confidentiality. The academic evolutionary computation community has apparently not devoted much energy to the car racing domain. One exception is Wloch and Bentley [11], who used evolutionary algorithms to optimize the parameters of simulated Formula 1 racing car with good results. However, they did not try to evolve the car controller, but rather used the simulator's built-in controller. Another interesting example is due to Floreano et al. [3], who evolved a controller for a first-person car racing game that successfully drove the car around the track. The controller used direct visual input to a neural network, but only a very small part (5 · 5 pixels) of the visual field was used at a time. This was achieved through letting the neural network select which part of the visual field to concentrate on by moving and focusing an artificial retina, a process known as active vision. Further, Stanley and Miikkulainen [7] evolved a collision warning system, but did not especially concern themselves with evolving good driving per se. At the same time as this work was done, Tanev et al. [8] evolved the parameters for a control algorithm for a physical model racing car. Worth mentioning also is the famous ALVINN experiments, where a neural network trained with back-propagation learned to keep a car on the road by observing a human driver's behaviour[6].

Applying evolutionary methods to car racing can be worthwhile from several perspectives. One is the development of appropriate automatic drivers for various purposes, such as more interesting/challenging racing game opponents that automatically can adapt to new tracks, or even racing physical cars of some sort. Another is automatically testing track/car combinations for "sweet spots", i.e. simple but non-desirable strategies a player can exploit to beat a computer game, something which is a great concern for the game industry [2]. But it is also worth considering car racing from an evolutionary robotics (ER) / embodied artificial intelligence perspective. Most ER experiments use robots with simple morphology and simple sensor setups in simple environments [5]. The archetypical ER research robot, the Khepera, can turn on the spot, accelerate and decelerate almost instantaneously and move with equal speeds forwards and backwards. These and other features set it apart from nearly every vehicle that is actually used for anything in the real world. At the same time, the sensor setup is very limited. While the availability of standardized research robots has doubtlessly benefited ER in many ways, it is plausible that the very limited dynamics, sensors, and environments used in such experimental setups is a major obstacle for ER to scale up, i.e. for evolution to produce really complex intelligence. We believe the more complex dynamics of the car racing problem could allow different and potentially more complex behaviour to evolve.

The main question we are trying to answer in this paper is what sort of information a control mechanism needs in order to proficiently race a car around a track, and how it should be represented. Along the way we will look at whether different input representations and control mechanisms give rise to qualitatively different driving styles, and not only quantitatively different lap times, and which representation/mechanism combinations give rise to general good driving behaviour as opposed to optimising behaviour for one particular way of driving around a track. (A problem here is the lack of objective measure driving quality.) Some constraints we adopt in these experiments are that all controllers are to be reactive (i.e. they do not learn or integrate information over time in any other way) and that we only use one track for evaluating them. Given these constraints, we implemented a wide range of simple evolvable control architectures.

### 1.1 Overview of the paper

After presenting the car model and the various algorithms we are using, we investigate two ways of evolving controllers that do not make use of any information about the car and the environment at all, but only about the elapsed time in the simulation. In control theory terminology, these are "open loop" controllers. One of these is based on neural networks and their oft-touted universal approximation capabilities. A neural network should in principle be able to approximate a function that maps a point in time to the optimal action at that time, though it is not known how easy such a function would be to evolve. The other is based on action sequences, which are lookup tables with one action per time step. An action sequence should be able to make an optimal controller, given a fixed starting point and a deterministic simulation, though there is a question of evolvability here as well. Also, minor variations in the starting position and angle of the car are likely to have detrimental effects on these controllers, something borne out by our experiments.

After that, we investigate two evolvable controllers which makes use of Newtonian information, i.e. "third-person" information about the state of the car, in this case velocity, position, orientation, angular velocity and speed. Again, the first of these controllers is based on a neural network, which here implements a function from the above mentioned inputs to car actions. The second is based on a force field approach, where the actual speed and orientation of the car is compared with the value associated with the particular part of the track the car occupies. This approach is broadly similar to that used in [4].

Finally, we investigate the use of neural networks to control the car based on egocentric information, i.e. "first person" data from simulated sensors whose characteristics are co-evolved with the neural networks.

## 2 The model

At the University of Essex, a permanent arena for racing miniature RC cars has been constructed, and similar arenas have been assembled for competition and demonstration sessions at the 2003 and 2004 IEEE Congresses on Evolutionary Computation (CEC), and a competition is also planned for CEC 2005. The arena measures approximately $2.4 \times 1.2$ metres, and the RC cars are 1/24th scale (approximately 17 centimeters long). The RC cars, which are of a type commercially available in toy stores, are controlled by the provided radio transmitter, attached to a PC by connecting the switches on the transmitter to a custom-made circuit board which plugs into the parallel port. The car controls are all on/off types (also known as bang-bang control), with the main motor either driving the car backward, forward or being un-powered, and the steering wheels pointing either to the left, to the right or towards the center, making nine possible control commands in all. Visual input is fed to the computer either from a web camera mounted on top of the track and connected via USB, or from a web camera mounted on top of the car and connected via a wireless link. Using this setup (with the overhead camera), hand-designed algorithms have driven the cars around the track, though with frequent collisions and intermittent progress.

For the experiments in this article, we have created a simple car-racing model aiming to reproduce many of the qualitative features of RC car racing. RC cars are much more robust than real cars (owing to their smaller size, smaller mass, and lower speed), and can bounce off walls rather than breaking down when crashing into them. We have not aimed to reproduce the characteristics of any particular make of RC car or surface material, however the shape of the track used in these experiments is modelled on the track used for the competition at CEC 2003. The dimensions of the model track is 400 pixels horizontal by 300 vertical, as depicted in Figure 1.

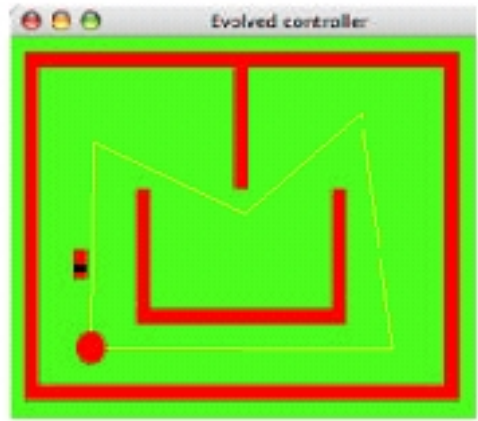

Figure 1: The simulated track with five aim points and the car at its starting position. The circle marks the aim point closest to the car.

Major differences between our current simulation and the real-world RC cars are as follows:

- In our simulation, the car position is known exactly at each point in time; in the real-world system, our computer vision system is imperfect, and while tracking the car fairly accurately most of the time, is occasionally prone to poor position estimates. Furthermore, our current real-world car tracker only estimates the position of the car, whereas the simulation also provides the orientation and heading of the car as possible inputs to a controller.

- The simulation proceeds via constant size time-steps, and has no latency (i.e., the *current* position of the car is available at each time step). The real system is subject to variable time delays, due to the computational demands of the vision system (the variability being a result of running on a multi-tasking operating system). Also, the real system often suffers a latency of over 100ms, which is time enough for the car to travel 30cm.

- We have implemented a simple model of skidding, whereby the car has some side-slip when cornering, which causes under-steer. On the real RC cars, this is highly non-linear, and can be exploited by skilled drivers to execute the equivalent of a hand-brake turn (which induces severe over-steer), to take very tight corners.

### 2.1 Dynamics

As computational efficiency is crucial in experiments in artificial evolution, the car simulation was designed to mini-mize the amount of computational effort needed per updating step, while still providing a qualitative similarity to the dynamics of real RC car racing. The simulation, which is implemented in Java, is based on Newtonian particle kinetics; the car's state is defined by its position, velocity, orientation and angular velocity.

The moving resistance of the car is modelled as a frictional force proportional to the velocity of the car. The traction of the tyres is limited, which causes skidding when cornering at high speed.

Wall collisions affect the velocity and angular velocity of the car in such a way that the car bounces off the walls in a realistic manner, and a collision at an unfortunate angle and speed can spin the car around so that it faces the wrong way, adding to the challenges of the task.

We did consider using an existing car racing simulation for our studies, such as RARS (Robot Auto Racing Simulation) [1], but found it easier to implement our own simulation in order to better match our requirements.

**2.2 Fitness function**

The fitness of a controller is determined by how far along the track it has managed to drive the car after a fixed amount of time; for the experiments in this paper this amount is specified as 500 time steps. How far the car has travelled is measured using a system of aim points. Five aim points are laid out quite evenly on the track, and the controller is scored on how many aim points it manages to reach in the correct order, which is counter-clockwise. To smooth the fitness function, the progress made towards the next aim point is also taken into account.

Each of the experiments below is carried out under two different regimes, a fixed starting point regime and a randomized starting point regime. Under the fixed starting point regime, the car always starts at pixel coordinates 50, 150, pointing straight downwards in the preferred direction of movement for that part of the track. Under the randomized starting point regime, the x-coordinate of the starting position is randomized between 40 and 60, and the y-coordinate between 120 and 180, while the orientation of the car is set to a random angle between $+/-$ 30 degrees from straight downwards. When calculating fitness under the randomized regime, each fitness evaluation is an average of three separate trials using different starting positions, in order to make the fitness function somewhat less noisy. Under the fixed regime, a very small random term is added to the fitness value in order to encourage neutral mutations.

The car can be controlled not only by an evolved controller, but also by a human player through a keyboard interface. When the car simulation was demonstrated at the IEEE 2005 Symposium on Computational Intelligence and Games, a number of conference delegates took the opportunity to try to beat the evolved controllers. Of the 10 participants who tried the game several times, the average best fitness (score) was 11.86. Overall winner was Jay Bradley of the University of Edinburgh, who once reached fitness 15.97.

**2.3 Evolutionary algorithm**

An evolutionary algorithm with truncation-based selection and elitism was used for all experiments. The workings of the algorithm were as follows: in each generation, the fitness of all individuals were evaluated, and the population was sorted in fitness order. The less fit half of the population was then deleted and replaced with a clone of the fitter half of the population, after which the mutation operator was applied to all individuals, except the two fittest individuals. All experiments in this paper used a population size of 100 and, unless otherwise stated, ran for 100 generations.

# 3 Neural networks

Three of the five experimental setups in this paper use neural network-based controllers. The neural networks are standard multi-layer perceptrons, with $n$ input neurons, a single layer of $h$ hidden neurons, and two output neurons, where each neuron implements the *tanh* transfer function. At each time step, the inputs as specified by the experimental setup is fed to the network, activations are propagated, and the outputs of the network are interpreted as actions that are used to control the car. Specifically, an activation of less than -0.3 of output 0 is interpreted as backward, more than 0.3 as forward and anything in between as no motor action; in the same fashion, activations of output 1 is interpreted as steering left, center or right.

At the start of an evolutionary run, the $m*n*2$ connections are initialized to strength 0. The mutation operator then works by applying a different random number, drawn from a gaussian distribution around zero with standard deviation 0.1, to the strength of each connection.

# 4 No inputs and action sequences

**4.1 Methods**

An action sequence is a one-dimensional array of length 500, containing actions, represented as integers in the range 0-8. An action is a combination of driving command (forward, backward, or neutral) and steering commands (left, right or center). When evaluating an action sequence controller, the car simulation at each time step executes the action specified at the corresponding index in the action sequence. At the beginning of each evolutionary run, controllers are initialized as sequences of zeroes. The mutation operator then works by selecting a random number of positions between 0 and 100, and changing the value of so many positions in the action sequence to a new randomly selected action.

**4.2 Results**

After evolving the action sequence controllers for 100 generations, most evolutionary runs reached a fitness of about 2 ; after 500 generations, they often reach about 5. The resulting behaviour looks more like rather random actions that just happen to take the car in the right direction, than it looks like good driving. The car drives very slowly, and many evolved controllers spend considerable amounts of time standing virtually still before finally starting to move. We hypothesize that a major factor restraining fitness growth is the ubiquity of local optima in the early parts of the sequence. This comes about because each action is associated with a time step rather than a position on the track, so that a mutation early in the sequence that is in itself beneficial (e.g. accelerating the car at the start of a straight track section) will offset the actions later in the sequence in such a way that it probably lowers the fitness as a whole, and is thus selected against.

---
[1] http://sourceforge.net/projects/rars

Under the randomized starting point regime, fitness is often below two and does not rise much further. Analysis of evolved controllers shows that the car often gets stuck on walls. A plot of fitness evolution for both the fixed and random starting points is shown in Figure 2.

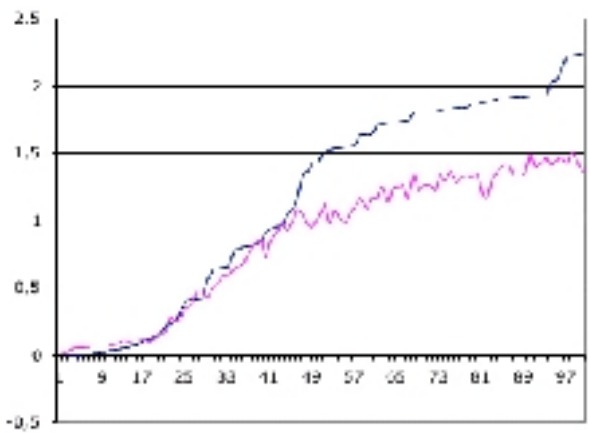

Figure 2: Evolving action sequences. The upper graph represents the fitness of the best individual in each generation, averaged over 10 evolutionary runs, under the fixed starting point regime. The lower graph represents the same entity when the car was evolved with randomized starting positions.

## 5 Open-loop neural network

### 5.1 Methods

A neural network as described above with two inputs, five hidden nodes and two outputs is fed with the number of the current time step divided by 500, yielding an input value of 0 in the first time step and 1 in the last, and a constant input with the value 1.

### 5.2 Results

After 100 generations of evolution, the controller typically reached fitness levels of about 2 to 3. The car behaviour looks no less random than that of the action sequence controllers, the main difference is that the car goes faster in this case. Most evolutionary runs found a way for the car to accelerate into the walls at the right angle and speed to bounce it's way around little more than half of the track, but none got further. An analysis of the actions produced by the controller reveals that the controller issues the same action for several hundred time steps in a row, and only changes action once or twice per trial. At the moment we don't know why any more intelligent behaviour refuses to evolve.

When evolving with randomized starting points it seems to be impossible to find a behaviour sequence that relies on bouncing off walls in the right way, and so evolved controllers tend to just run around in circles and reach very low fitness levels, barely above zero. The fitness evolution of this type of controller is shown in Figure 3.

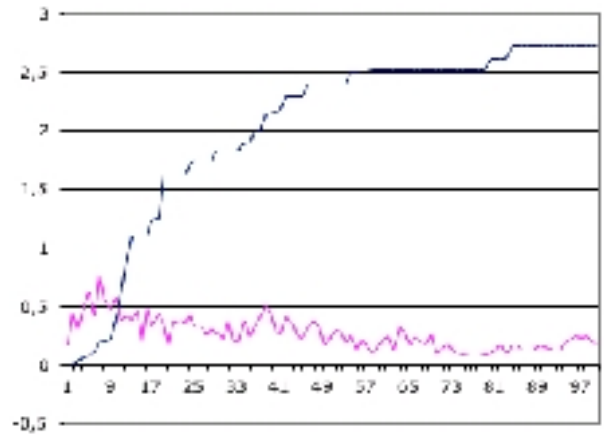

Figure 3: Evolving open loop neural network controllers. The upper line is for a fixed starting position, the lower line for randomised starts.

## 6 Newtonian inputs and force fields

### 6.1 Methods

A force field controller is here defined as a two-dimensional array of two-tuples, describing the preferred speed and preferred orientation of the car while it is in the field. Each field covers an area of $n*n$ pixels, and as the fields completely tile the track without overlapping, the number of fields are $(l/n)*(w/n)$, where $l$ is length, and $w$ is width of the track, respectively. At each time-step, the controller finds out which field the centre of the car is inside, and compares the preferred speed and orientation of that field with the cars actual speed and orientation. If the actual speed is less than the preferred, the controller issues an action containing a forward command, otherwise it issues a backward command; if the actual orientation of the car is left of the preferred orientation, the issued action contains a steer right command, otherwise it steers left. In the results reported here, we used fields with the size $20 \cdot 20$ pixels, evolved with gaussian mutation with magnitude 0.1, though we have tried other combinations of field size and mutation magnitude without any improvement in fitness. This is broadly similar to the kind of controllers evolved in [4], though we are controlling a car rather than a holonomic robot.

### 6.2 Results

The force field controllers evolve very slowly, and after 100 generations barely exceeded fitness 1; evolving for 1000 generations sometimes brought fitness up to around 4 when using fixed starting positions; when starting positions were randomised, fitness stayed at 1. The cars moved around in a peculiar fashion, sometimes following a sane path around the track for a while, only to become stuck oscillating between two force fields a moment later. Figure 4 shows the fitness evolution of this type of controller, and Figure 5 shows a sample trace.

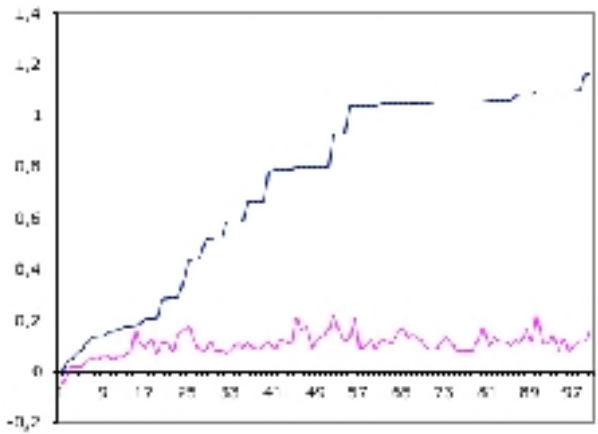

Figure 4: Evolving force field controllers.

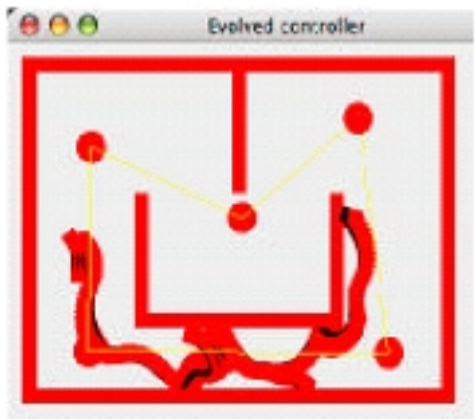

Figure 5: Movement trace of a car controlled by a force field controller.

## 7 Newtonian inputs and neural networks

### 7.1 Methods

The neural network is fed seven inputs: a constant input with value 1, the *x* and *y* components of the car's position, the *x* and *y* components of its velocity, its speed and its orientation. All inputs are scaled to be in the range -10 to 10. Seven hidden neurons are used in the network, and the two outputs are interpreted as described above.

### 7.2 Results

Evolving for 100 generations, best fitness varies considerable between evolutionary runs. While most runs produced controllers with fitness values around 3, at least one run produced a controller with fitness over 6. None of the controllers manage to the drive the car properly around the track however, the fittest controller instead drove the car into the "box" in the center of the track and exploited a glitch in the fitness function, whereby it can come close enough to the aim points on the left side of the track for the fitness function to increase without the car ever going around the left wall of the box. The cars drive fast and seem to make sensible turns, but they all eventually get stuck on a wall.

Randomising the starting position produces controllers of slightly lower fitness. Figure 6 shows the fitness evolution of this type of controller, while Figure 7 shows a sample trace of the car.

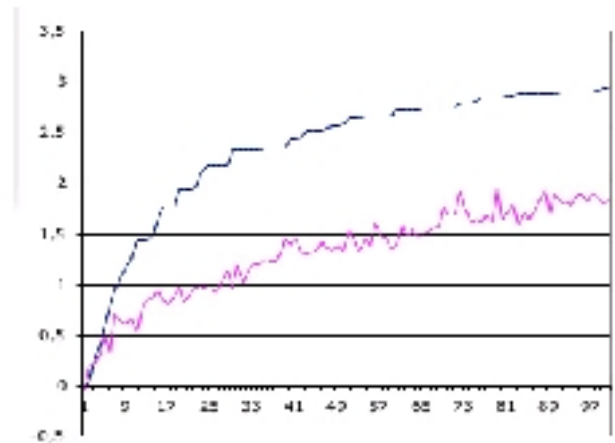

Figure 6: Evolving newtonian neural network controllers.

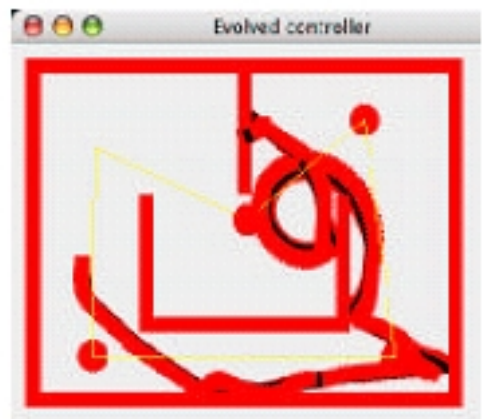

Figure 7: Movement trace of a car controlled by a neural controller with newtonian inputs.

# 8 Simulated sensor inputs and neural networks

## 8.1 Methods

In this experimental setup, the six inputs to the neural network consist of one constant input with the value 1, the speed of the car, and the outputs of three wall sensors and one aim point sensor. The aim point sensor simply outputs the difference between the car's orientation and the angle from the center of the car to the next aim point, yielding a negative value if that point is to the left of the car's orientation and a positive value otherwise.

Each of the three wall sensors is allowed any forward facing angle (i.e. a range of 180 degrees), and a reach between 0 and 100 pixels. These parameters are co-evolved with the neural network of the controller. The sensor works by checking whether a straight line extending from the centre in the car in the angle specified by that sensor intersects with the wall at eleven points positioned evenly along the reach of the sensor, and returning a value equal to 1 divided by the position along the line which first intersects a wall. Thus, a sensor with shorter reach has higher resolution, and evolution has an incentive to optimize both reaches and angles of sensors. This type of sensor controller is related to the wrap-around vector histogram approach of [1], except that we are only using three sensors instead of full wrap-around.

## 8.2 Results

After 100 generations, evolution produced a controller with excellent fitness values, which equal more than three laps around the track in the allotted 500 time steps. The cars drive around the track at close to full speed, cutting corners incredibly close, crashing into walls only where they can take advantage of the rebound. The sensors vary considerably in the combination of angles and ranges, though often show a bias towards straight ahead and left. An example evolved sensor configuration is shown in Figure 8, which uses the short left sensor to help follow the inside wall, or take close cut corners, and the longer range sensors to help decide when to turn.

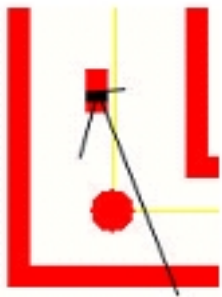

Figure 8: A sample evolved configuration of sensors.

The best final fitness value found when examining ten evolved controllers was 16.04, which narrowly beats the best human competitor so far. When evolving with randomised starting positions best fitness was slightly lower, with a similar sensor setup. The lower fitness is due to the cars slowing down in corners. Figure 9 shows the fitness evolution of this type of controller, while Figure 10 shows a sample trace of the car's movement.

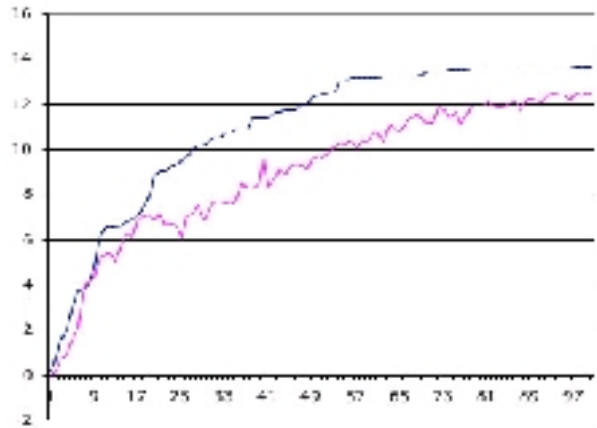

Figure 9: Evolving sensor-based neural controllers with full inputs.

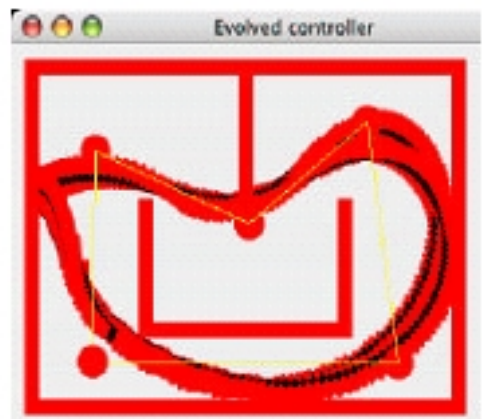

Figure 10: Movement trace of a car controlled by a sensor controller.

To investigate the relative contributions of the wall and aim point sensors, we "lesioned" the controller by disabling the sensor types one at a time. First, we disabled the wall sensors, and evolved controllers making use only of the aim point sensor, speed and the constant input. Under the fixed starting point regime this resulted in cars with fitness often between 11 and 12; they drove well, but bumped into the wall protruding from the top of the track once every lap. When randomizing starting points, the aim sensor-only controller fared much worse, reaching medium fitness about 7. Evolution produced a controller that sometimes made its way around the track, but, depending on initial conditions, more often got stuck on a wall. Figure 11 shows the fitness evolution under this restriction.

We then re-enabled the wall sensors and instead disabled the aim point sensor. Under the fixed starting point regime, evolution produced controllers that often had all wall sensors set long range, and pointing approximately 20 degrees left of straight ahead, and that drove at high speed, more or less following the outer wall. When randomizing starting points, they often have a long range sensor pointing straight forward, and medium range sensors pointing approximately 45 and 90 degrees to the left, and execute careful following of the outer wall without ever bumping into it.

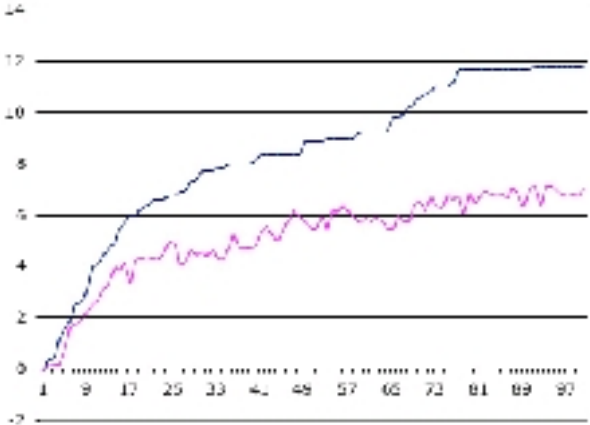

Figure 11: Evolving sensor-based neural controllers without wall sensors.

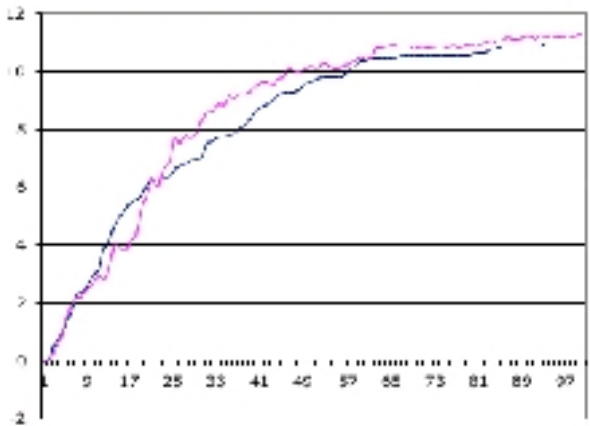

Figure 12: Evolving sensor-based neural controllers without aim point sensors. Note that the value for the random starting point regime is actually slightly higher after generation 25.

| Controller | Fixed | Randomized |
|---|---|---|
| Action sequence | 2.23 | 1.36 |
| Open loop neural | 2.72 | 0.17 |
| Force field | 1.16 | 0.16 |
| Newtonian neural | 2.94 | 1.84 |
| Sensor-based | 13.59 | 12.4 |
| No wall sensors | 11.76 | 7.02 |
| No aim point sensor | 10.97 | 11.33 |

Table 1: Average fitness of best individuals of 10 evolutionary runs of the various controller architectures under the starting point regimes.

### 8.3 Time consumption

Evolving 100 generations of a population of 100 sensor-based controllers, where each controller is evaluated for 500 time steps, takes 4 minutes and 22 seconds using Java 1.4.2 on a 933 Mhz Apple iBook G4. This works out to approximately 2.82 seconds per generation, 28 ms per controller evaluation, or 0.05 ms per time-step. A little more than half of this time is spent on propagating activations through the neural network. While the evolved networks should be fast enough to control many cars simultaneously in a commercial driving simulator, the process of evolving the controllers is probably still too slow to e.g. create new drivers adapted for arbitrary tracks on the fly.

## 9 Conclusions

The most consistent effect across all the experiments reported above is that controllers (except the sensor controller with the aim point sensor disabled) have higher fitness when starting position and orientation is kept fixed. Not surprisingly, evolution is able to optimize car behaviour better in these noise-free cases, and has to develop more robust behaviour (which means longer lap times) or risk getting stuck on a wall when starting position is randomized. When racing actual physical cars, starting position will necessarily vary, and so performance under this regime is the more interesting factor when evaluating the suitability of a controller for transfer to a physical domain.

Our experiments also point to the vast superiority of first-person to third-person information for the problem at hand. This might be because of the existence of walls, which presumably makes any mapping from third-person spatial information to appropriate first-person actions, extremely nonlinear. A controller using third-person information (such as visual data from an overhead web camera) could get around this problem by somehow representing the walls, and re-creating the kind of sensors used in our sensor-based simulations described above. The problems with using third-person information might also partly be due to difficulty of rotating coordinates as the car's orientation changes.

We were somewhat surprised by the poor performances of the action sequence and force field controllers, both of which should theoretically be able to represent good solutions, at least for fixed starting points. We therefore hypothesize that the poor performance is because of problems with the evolutionary algorithm rather than the representations per se; our main culprit here is the mutation methods, which seem to drive the action sequence into local optima and make for very slow progress in force field evolution. For a fixed starting position, it should be possible to achieve good lap times by seeding the EA with an action section observed by running an evolved sensor controller, but we've not yet tried this.

Regarding force field controllers, an alternative hypothesis is that it is simply not theoretically possible to successfully drive a non-holonomic vehicle around a track using that paradigm directly, as it does not take into account the state of the car when entering a particular cell (this is not a problem for holonomic vehicles, which can be treated as stateless).

**9.1 Future research**

To investigate whether the evolved controllers really display good driving behaviour as opposed to optimizations for one particular track, we would need to test them on more than one track. This could be done by testing each individual on several different track in each fitness evaluation, and then using a different set of tracks for testing the final evolved driving ability. Possibly a "generalist" driver could be evolved that performs reasonably on all tracks; this controller could then, using incremental or layered evolution [9], be specialized to perform well on a particular track — hopefully in shorter time than it would take to evolve a specialized controller from scratch.

The neural networks in these experiments are quite simple; recurrent neural networks, or even learning, plastic networks could improve track times and generality of driving skills further. Several times, the Newtonian input neurocontroller evolved to exploit certain weak spots in the fitness function; these must be rectified.

So far we have only evolved controllers for a time-trial task, involving driving a single car at maximum speed. There is much work to be done, however, on competitive racing, involving two or more cars on the track simultaneously, with interesting adversarial collision strategies to be evolved. An interesting difference between this and other forms of competitive coevolution is the existence of an absolute, and not just a relative fitness function.

While the car model used in this paper is adequate for the experiments mentioned above, we plan to investigate the application of our evolutionary methods to commercial driving simulators and physical radio-controlled cars. An important research question for the latter context is how the evolutionary methods handle the time-lag in the physical control cycle, as reported by Tanev et al. [8]

But the most challenging and interesting future development is using raw visual data as input to the controllers. Visual input for controlling a first-person car game has already been explored by Floreano et al., but much work certainly remains to be done here. In particular, Floreano et al.'s controllers made use only of a minuscule part of the visual field; we aim to use convoluted modular neural networks [10] as the basis for controllers that make use of raw visual data, from a first person or a third person perspective, for controlling both physical and simulated cars. Currently, we are working on evolving controllers for the simulated car whose input is the 400 ∗ 300 track image rotated and translated so that the car is always in the centre of the picture and facing upwards. There are some indications that it is possible to train perceptrons to control the car based on such input data.